# Unsupervised Detection of Fraudulent Transactions in E-commerce Using Contrastive Learning

Xuan Li
Columbia University
New York, USA

Yuting Peng
New York University
New York, USA

Xiaoxuan Sun
Independent Researcher
Mountain View, USA

Yifei Duan
University of Pennsylvania
Philadelphia, USA

Zhou Fang
Georgia Institute of Technology
Atlanta, USA

Tengda Tang *
University of Michigan
Ann Arbor, USA

*Abstract*-With the rapid development of e-commerce, e-commerce platforms are facing an increasing number of fraud threats. Effectively identifying and preventing these fraudulent activities has become a critical research problem. Traditional fraud detection methods typically rely on supervised learning, which requires large amounts of labeled data. However, such data is often difficult to obtain, and the continuous evolution of fraudulent activities further reduces the adaptability and effectiveness of traditional methods. To address this issue, this study proposes an unsupervised e-commerce fraud detection algorithm based on SimCLR. The algorithm leverages the contrastive learning framework to effectively detect fraud by learning the underlying representations of transaction data in an unlabeled setting. Experimental results on the eBay platform dataset show that the proposed algorithm outperforms traditional unsupervised methods such as K-means, Isolation Forest, and Autoencoders in terms of accuracy, precision, recall, and F1 score, demonstrating strong fraud detection capabilities. The results confirm that the SimCLR-based unsupervised fraud detection method has broad application prospects in e-commerce platform security, improving both detection accuracy and robustness. In the future, with the increasing scale and diversity of datasets, the model's performance will continue to improve, and it could be integrated with real-time monitoring systems to provide more efficient security for e-commerce platforms.

## I. INTRODUCTION

With the rapid development of e-commerce, online shopping has become one of the main consumption methods for global consumers. According to statistical data, the global e-commerce market continues to expand, especially with the boost of mobile internet [1]. This has led to profound changes in consumer purchasing behavior. These changes not only bring convenience to consumers but also present enormous business opportunities for merchants [2]. However, the rise of e-commerce fraud has gradually become a serious security threat, impacting merchants' profits and consumers' trust. E-commerce fraud includes a variety of behaviors, such as account theft, fake transactions, payment fraud, and malicious returns [3]. In severe cases, it can lead to loss of merchant funds and damage to brand image. Therefore, accurately and quickly identifying and preventing e-commerce fraud has become one of the core tasks in e-commerce platform security systems.

At present, traditional e-commerce fraud detection methods mostly rely on manual feature engineering and rule-based models. These methods typically require manually defining numerous features and setting rules based on expert experience. However, this approach often struggles to address the ever-evolving and diverse nature of fraud, as new fraudulent patterns emerge continuously. The maintenance and updating of manual rules are also challenging. Moreover, manual feature engineering requires significant time and effort but cannot fully uncover complex patterns hidden in the data. Consequently, an increasing number of research endeavors are adopting machine learning and deep learning methodologies to address this challenge. Deep learning can automatically extract features and model complex relationships in data, making it better suited to the diversity and variability of fraudulent behaviors [4].

In the artificial intelligence domain, contrastive learning has recently become an important unsupervised learning method. Unlike traditional supervised learning methods, contrastive learning does not rely on labeled data [5]. Instead, it learns representations by constructing positive and negative sample pairs and has been widely adopted in the medical field [6-7]. By maximizing the similarity between samples of the same class and minimizing the distance between samples of different classes, contrastive learning can learn robust and generalizable feature representations. Therefore, applying contrastive learning methods for the unsupervised detection of e-commerce fraud allows for fraud detection without labeled data. It can identify fraudulent behaviors by learning the intrinsic relationships between samples. Compared to traditional rule-based and manual feature methods, contrastive learning-based models can automatically discover hidden fraud patterns in data, improving the model's adaptability and accuracy [8].

SimCLR (Simple Contrastive Learning of Representations), a popular contrastive learning method, has achieved significant success in fields such as image and text processing [9], but remains underexplored in e-commerce fraud detection. This study proposes a SimCLR-based unsupervised algorithm to identify fraudulent transactions by learning hidden patterns from unlabeled data. By constructing positive and negative sample pairs, the model enhances fraud differentiation while reducing reliance on labeled datasets. This approach improves detection accuracy and efficiency, addressing the growing sophistication of fraud. As fraudulent activities evolve, leveraging contrastive learning offers a promising solution to strengthen e-commerce security with minimal labeling costs.

## II. RELATED WORK

With the growing complexity of fraudulent behaviors in e-commerce transactions, traditional rule-based and supervised learning methods face increasing limitations in adaptability and scalability. To address these challenges, recent studies have explored deep learning techniques, self-supervised learning, and statistical modeling for fraud detection and anomaly identification.

Deep learning models have demonstrated strong capabilities in capturing complex patterns within sequential and high-dimensional data. Feng [10] introduced a hybrid approach that integrates bidirectional long short-term memory (BiLSTM) with self-attention mechanisms, effectively improving pattern recognition in sequential datasets. Similarly, Deng [11] proposed a model that leverages deep networks to enhance the understanding of temporal dependencies, reinforcing the effectiveness of automated feature extraction. These approaches suggest that deep learning architectures can play a crucial role in detecting fraudulent activities without the need for extensive manual feature engineering. Self-supervised learning has recently emerged as a powerful technique for representation learning, eliminating the need for labeled datasets. Wei et al. [12] explored a self-supervised framework that enhances feature extraction in complex data environments, demonstrating how structured relationships between data points can improve learning efficiency. Liao et al. [13] further refined this approach by integrating knowledge representations to optimize learning in ambiguous and dynamic settings, highlighting the potential of self-supervised methods to uncover hidden patterns in large-scale data.

Contrastive learning, particularly SimCLR, has gained significant attention for its ability to learn meaningful representations in an unsupervised manner. Gao et al. [14] demonstrated the effectiveness of meta-learning and contrastive loss functions in learning generalizable representations from limited data, reinforcing the applicability of contrastive learning to fraud detection. These insights align with the objective of this study, which seeks to identify fraudulent patterns by distinguishing meaningful representations within unlabeled transactional data. Beyond neural network-based models, statistical and data-driven approaches have also contributed to addressing challenges in fraud detection. Wang [15] introduced an adaptive classification framework that improves learning from imbalanced datasets, a common issue in fraud detection. Additionally, Wang also [16] proposed an optimized data fusion method that enhances robustness against inconsistencies in transaction records, showcasing the importance of integrating multiple data perspectives to improve detection accuracy. These methodologies contribute valuable insights into refining fraud detection strategies by optimizing feature selection and improving model generalization.

Other research efforts have focused on refining representation learning and anomaly detection techniques. Yan et al. [17] explored methods for extracting structured information from complex sequences, providing a foundation for improving fraud detection in evolving transactional data. Studies such as Liu [18] and Zhou et al. [19] demonstrated how leveraging multi-source information and structured transformations can enhance decision-making, further supporting the role of adaptive learning models in fraud detection scenarios. Additionally, Zhan [20] examined methods to enhance system adaptability, reinforcing the need for dynamic learning models capable of handling real-time variations in data distributions.

## III. METHOD

In this study, we proposed an unsupervised e-commerce fraud detection algorithm based on SimCLR, which uses a contrastive learning mechanism to learn the latent representation of transaction data by constructing positive and negative sample pairs [21]. To achieve this goal, we first defined the input features of the transaction data and input them into the SimCLR model for training to automatically learn the representation of fraudulent behavior. SimCLR uses an optimization method based on a contrastive loss function to maximize the similarity between positive sample pairs and minimize the distance between negative sample pairs, thereby achieving efficient unsupervised learning. The core parts of the method, including the design of the model architecture and loss function, are introduced in detail below. Its model architecture is shown in Figure 1.

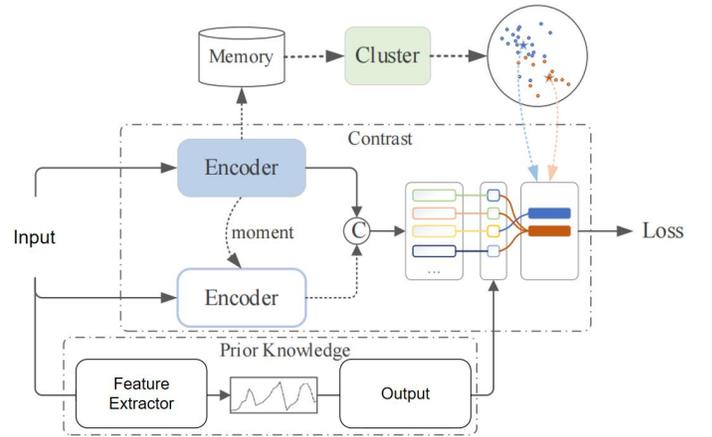

Figure 1 Model network architecture

First, for the processing of e-commerce transaction data, we consider each transaction record as a sample, which includes multiple features, such as transaction amount, type of purchased goods, transaction time, etc. In practical applications, e-commerce transaction data is usually high-dimensional and contains multiple categories of features [22]. In order to make the data suitable for the input of the deep learning model, we standardize the original data and adjust the mean of each feature to 0 and the variance to 1. The mathematical representation of standardizing feature data is as follows:

$$x_i^{'} = \frac{x_i - \mu_i}{\sigma_i}$$

Among them $x_i$ is the i-th feature, $\mu_i$ is the mean of the i-th feature, $\sigma_i$ is the standard deviation of the i-th feature, and $x_i^{'}$ is the standardized feature value.

After preprocessing the data, we entered the training phase of the model. The core idea of SimCLR is to learn the representation of data through contrastive loss functions so that similar samples are close in the latent space and different samples are far away [23]. In order to construct positive and negative sample pairs, we perform random data augmentation operations on the e-commerce transaction data, including perturbing the original transaction data (such as adding noise, randomly changing the transaction amount, etc.). We assume that each transaction record can obtain two views through different augmentation methods, denoted as $x_i^A$ and $x_i^B$. Then, we use a deep neural network $f(\cdot)$ to map each view to a representation in the latent space:

$$h_i^A = f(x_i^A), h_i^B = f(x_i^B)$$

Where $h_i^A$ and $h_i^B$ are the representations of the two views of sample $i$ in the latent space. Next, the contrastive loss function is used to optimize these representations so that the two views of the same sample are as close as possible and the representations of different samples are as far apart as possible. The contrastive loss function used in SimCLR is usually based on cosine similarity to measure the similarity between two samples. For any pair of samples $i$ and $j$, the cosine similarity is defined as:

$$sim(h_i, h_j) = \frac{h_i^T h_j}{\|h_i\| \|h_j\|}$$

To optimize the contrastive learning model, we use a contrastive loss function, which is:

$$L_{con} = -\log \frac{\exp(sim(h_i^A, h_i^B)/\tau)}{\sum_{k=1}^{N} \exp(sim(h_k^A, h_k^B)/\tau)}$$

Where $\tau$ is the temperature hyperparameter and $N$ is the total number of samples in the training set. The purpose of this loss function is to maximize the similarity between two views of the same transaction and to distance different samples through negative samples. When optimizing this loss function, we update the parameters of the network through the gradient descent method and finally obtain the potential representation of the e-commerce transaction data [24].

After training the contrastive learning model, we obtain the latent representation $h_i$ of each transaction sample. These latent representations not only contain the main features of each transaction but also reflect the similarities and differences between samples. In the fraud detection task, we need to judge whether the transaction is a fraudulent transaction based on these representations. To this end, we first calculate the similarity between each sample and all other samples and determine whether the sample is abnormal (i.e., fraudulent behavior) based on these similarities. Specifically, for a transaction sample $x_i$ to be detected, we calculate the cosine similarity between its latent representation $h_i$ and the latent representation $h_j$ of all other samples:

$$sim(h_i, h_j) = \frac{h_i^T h_j}{\|h_i\| \|h_j\|}$$

Next, we perform anomaly detection based on the distribution of similarity. If the similarity of sample $h_i$ is lower than that of most other samples, that is, its potential representation is significantly different from that of other normal transactions, then the transaction is considered abnormal (fraudulent behavior). To quantify this difference, we can calculate the average similarity and standard deviation of sample $x_i$, and then set a threshold to determine whether it is fraudulent behavior. Specifically, the average similarity is:

$$\mu_i = \frac{1}{N-1} \sum_{j \neq i} sim(h_i, h_j)$$

The standard deviation is:

$$\sigma_i = \sqrt{\frac{1}{N-1} \sum_{j \neq i} (sim(h_i, h_j) - \mu_i)^2}$$

Based on the average similarity and standard deviation, we can set a threshold t to determine whether sample $x_i$ is abnormal. If $\mu_i - \sigma_i$, then sample $x_i$ is determined to be a fraudulent transaction.

In summary, this method uses the contrastive learning algorithm SimCLR to unsupervisedly learn the potential representation of e-commerce transaction data and combines similarity measurement to identify fraudulent behavior. The whole process avoids the reliance on labeled data in traditional methods, can automatically discover and identify fraudulent behavior in e-commerce platforms, and provides an effective security protection mechanism for e-commerce platforms.

IV. EXPERIMENT

A. Datasets

This study utilizes publicly available transaction data from eBay, one of the largest global e-commerce platforms, which offers a diverse and extensive dataset covering various product categories. eBay's transaction data is well-suited for fraud detection research as it includes both normal and potentially fraudulent transactions, such as fake listings, payment fraud, and account takeovers. The dataset, characterized by complex user behaviors and transaction patterns, provides rich material for unsupervised learning. After preprocessing, irrelevant data was removed, retaining key features such as transaction amount, buyer and seller details, transaction status, and product category. The prevalence of unlabeled fraudulent transactions makes this dataset ideal for training a contrastive learning-based model in an unsupervised setting. Furthermore, its coverage of diverse fraud types enhances the practical applicability of this research in improving e-commerce security.

*B. Experimental Results*

To validate the effectiveness of the proposed SimCLR-based unsupervised e-commerce fraud detection algorithm, comparative experiments were conducted against traditional unsupervised learning methods. The evaluation metrics included accuracy, precision, recall, and F1 score, with K-means, Isolation Forest, Autoencoders, and Variational Autoencoders selected as baseline models. These experiments aimed to assess the fraud detection capability of SimCLR in an unlabeled data environment. The results, presented in Table 1, demonstrate the performance differences between the proposed method and traditional approaches.

Table 1 Experimental results

| Model | AUC | Precision | Recall | F1-Score |
|---|---|---|---|---|
| K-means | 0.8521 | 0.8145 | 0.7583 | 0.7850 |
| Isolation Forest | 0.8883 | 0.8387 | 0.8024 | 0.8203 |
| Autoencoders | 0.9035 | 0.8541 | 0.8321 | 0.8429 |
| Variational Autoencoders | 0.9426 | 0.8654 | 0.8498 | 0.8575 |
| Ours | 0.9849 | 0.9321 | 0.8737 | 0.9177 |

The experimental results demonstrate that the SimCLR method significantly outperforms traditional unsupervised learning models, particularly in AUC, precision, recall, and F1 score. Compared to K-means, Isolation Forest, Autoencoders, and Variational Autoencoders, SimCLR achieved an AUC of 0.9849, highlighting its superior ability to distinguish between normal and fraudulent transactions. While Autoencoders and Variational Autoencoders perform well in AUC and precision, their lower recall and F1 scores indicate limitations in detecting fraud. K-means and Isolation Forest, though effective in certain cases, fail to capture complex data features as effectively as SimCLR. These findings underscore the potential of contrastive learning for improving unsupervised fraud detection. Figure 2 presents the AUROC curve, further illustrating the model's performance.

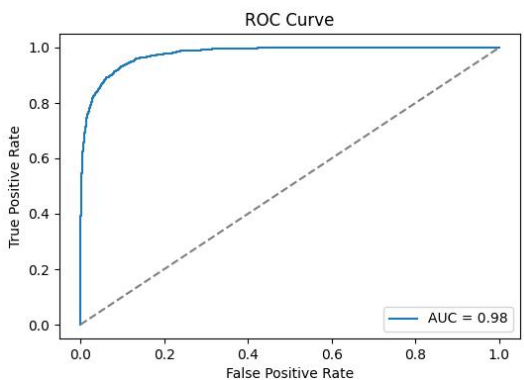

Figure 2 ROC curve

As can be seen from the ROC curve in the figure, the model performs very well, with an AUC value of 0.98, close to 1, which means that the model is very effective in distinguishing fraud from normal transactions. The ROC curve shows a clear upward trend, indicating that the model can maintain a high true positive rate under different false positive rates, which shows that the model has strong detection capabilities. Generally speaking, the closer the AUC value is to 1, the better the model's discrimination ability is. An AUC value of 0.98 indicates that the model's discrimination ability in fraud detection has almost reached the optimal level.

In addition, the curve is close to the upper left corner, further verifying that the model can effectively reduce the false positive rate and false negative rate, which is crucial for e-commerce fraud detection. In practical applications, a higher AUC value usually means that the model can accurately distinguish normal transactions from fraudulent transactions, reduce the risk of misjudgment, and thus improve the security of the platform. Overall, this ROC curve shows the superior performance of the model in distinguishing e-commerce fraud in an unsupervised learning environment.

Finally, this paper also gives a confusion matrix diagram, as shown in Figure 3.

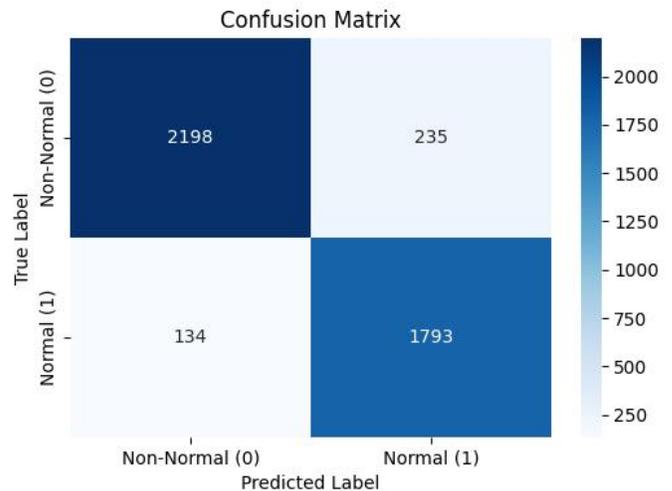

Figure 3 Confusion matrix diagram of experimental results

From the confusion matrix, it can be seen that the model performs well in predicting both normal and abnormal transactions. For abnormal transactions (i.e., fraud transactions), the model successfully predicted 2,198 fraudulent transactions (true positives), while 235 fraudulent transactions were incorrectly predicted as normal transactions (false negatives). For normal transactions, the model accurately predicted 1,793 normal transactions (true positives), but 134 normal transactions were misclassified as fraudulent transactions (false positives). This indicates that, despite a small number of false positives and false negatives, the model performs quite well and is able to effectively distinguish between normal and fraudulent transactions.

Overall, the model's performance is balanced, with relatively low false positives and false negatives, especially excelling in recognizing normal transactions. The number of mispredictions is relatively small, which suggests that the model has high reliability in e-commerce fraud detection. Although there are some false positives and false negatives, these errors are acceptable for complex fraud detection tasks and do not significantly affect the overall performance of the system.

## V. CONCLUSION

This study proposes an unsupervised e-commerce fraud detection algorithm based on SimCLR. Using contrastive learning, the model can effectively identify fraudulent activities in e-commerce transactions in an unlabeled data environment. The experimental results show that the proposed algorithm significantly outperforms traditional unsupervised learning methods, such as K-means, Isolation Forest, and Autoencoders, particularly in metrics like AUC, precision, recall, and F1 score. By learning the underlying representations of transaction data, SimCLR can automatically capture the differences between normal and fraudulent transactions, thereby improving the fraud detection capabilities of e-commerce platforms.

However, despite the good performance of the proposed algorithm on the current dataset, certain limitations still exist. Due to the diversity and complexity of e-commerce fraud behaviors, the model may experience some misclassifications or missed detections when dealing with different types of fraud. Furthermore, the scale and diversity of the dataset significantly impact the model's performance. Future research could explore the use of larger and more representative datasets to improve the model's generalization ability and robustness. Looking ahead, unsupervised fraud detection methods based on contrastive learning have significant development potential. With the continuous advancement of deep learning techniques, the efficiency and accuracy of models in handling large-scale e-commerce transaction data will improve. Additionally, incorporating more advanced feature engineering and data augmentation techniques may further enhance the model's performance. In terms of application, the algorithm could be integrated with real-time transaction monitoring systems for more efficient fraud detection. As multimodal data integration advances, the model's performance will continue to optimize, providing e-commerce platforms with smarter and more accurate security solutions